\documentclass[11pt,a4paper]{llncs}
\usepackage{amsmath}
\usepackage{amssymb}
\usepackage{graphicx}
\usepackage{marvosym}
\usepackage{url}
\usepackage{fancyhdr}
\usepackage[colorlinks=true, allcolors=blue]{hyperref}

\usepackage{geometry}
\newcommand{\keywords}[1]{\par\addvspace\baselineskip
\noindent\keywordname\enspace\ignorespaces#1}

\fancypagestyle{firstpage}{\fancyhf{}
}

\begin{document}


\title{\LARGE{Heterogeneous Entity Matching with Complex Attribute Associations using BERT and Neural Networks}}


%
%
\author{\large{Jiamin.Lu  \and Shitao. Wang }}
\institute{\large{Key Laboratory of Water Big Data Technology of Ministry of Water Resources\\Hohai University, Nanjing, China\\211307040019@hhu.edu.cn }}

%


%
%


\maketitle

\thispagestyle{firstpage}

\begin{abstract}
Across various domains, data from different sources such as Baidu Baike and Wikipedia often manifest in distinct forms. Current entity matching methodologies predominantly focus on homogeneous data, characterized by attributes that share the same structure and concise attribute values. However, this orientation poses challenges in handling data with diverse formats. Moreover, prevailing approaches aggregate the similarity of attribute values between corresponding attributes to ascertain entity similarity. Yet, they often overlook the intricate interrelationships between attributes, where one attribute may have multiple associations. The simplistic approach of pairwise attribute comparison fails to harness the wealth of information encapsulated within entities.To address these challenges, we introduce a novel entity matching model, dubbed "Entity Matching Model for Capturing Complex Attribute Relationships (EMM-CCAR)," built upon pre-trained models. Specifically, this model transforms the matching task into a sequence matching problem to mitigate the impact of varying data formats. Moreover, by introducing attention mechanisms, it identifies complex relationships between attributes, emphasizing the degree of matching among multiple attributes rather than one-to-one correspondences. Through the integration of the EMM-CCAR model, we adeptly surmount the challenges posed by data heterogeneity and intricate attribute interdependencies. In comparison with the prevalent DER-SSM and Ditto approaches, our model achieves improvements of approximately 4\% and 1\% in F1 scores, respectively. This furnishes a robust solution for addressing the intricacies of attribute complexity in entity matching.
\end{abstract}


\section{Introduction}

Knowledge graph update\cite{hogan2021knowledge} is a dynamic process of maintaining and revising existing knowledge graphs to reflect the ever-changing landscape of real-world knowledge. In this context, entity matching (EM) assumes paramount importance as different data sources continuously evolve, leading to a more complex and challenging knowledge graph update.

Entity Matching (EM)\cite{li2021deep} aims to determine whether different data references point to the same real-world entity. The objective of EM is to ascertain if data belongs to the same hydraulic entity. In entity matching, data can be classified into two categories\cite{agarwal2023named}: homogenous data and heterogeneous data. Homogenous data refers to data with the same schema, meaning they share identical attribute names. Based on the correctness of attribute values and their alignment with attribute names, homogenous data can be further categorized into clean data and dirty data. Clean data indicates that attribute values are correctly placed under the appropriate attributes, i.e., attribute values are aligned with corresponding attributes in the schema. Dirty data \cite{brunner2020entity} implies that attribute values might be erroneously placed under the wrong attributes, i.e., attribute values are not aligned with corresponding attributes in the schema. Heterogeneous data, on the other hand, involves dissimilar attribute names and may exhibit one-to-one, one-to-many, or many-to-many correspondence relationships.

Entity matching \cite{christophides2020overview} typically involves two steps: blocking\cite{li2020survey} and matching\cite{christophides2019end}. The purpose of blocking is to reduce computational costs by partitioning records into multiple blocks, where only records within the same block are considered potential matches. Subsequently, within each block, matching is performed to identify valid pairs of matching records, which is a crucial step in the entity matching process. However, in the matching process, prevalent models often encounter attribute matching issues, particularly when dealing with heterogeneous data.

\begin{figure} [ht]
   \begin{center}
   \begin{tabular}{c} 
   \includegraphics[height=5cm]{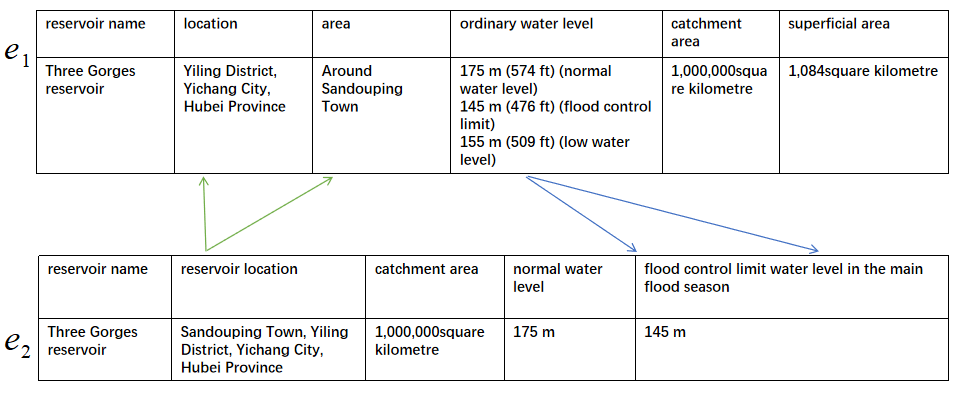}
   \end{tabular}
   \end{center}
   \caption[example] 
   { \label{fig:example} 
Wikipedia and Baidu Baike information about the Three Gorges Reservoir.}
   \end{figure}

As these models \cite{peeters2021dual} \cite{thirumuruganathan2018reuse}  \cite{teong2020schema} typically concentrate on homogenous data (often directly performing entity matching on structured database tables), they neglect the consideration of heterogeneous data (i.e., data scraped from web pages, where data attributes exhibit substantial variations). As depicted in Fig.\ref{fig:example},$e_1$ and $e_2$ respectively represent heterogeneous data extracted from Wikipedia and Baidu Baike about the Three Gorges Reservoir. Their attribute names are not identical and complex correspondence relationships exist.For candidate entity pairs $(e_1,e_2)$,conventional EM methods tend to compare tokens based on properties like "Location" and "Reservoir Location," due to their highest token similarity. However, the attribute "Reservoir Location" encompasses information related to both "Location" and "Region," and a simplistic token-based similarity assessment between "Reservoir Location" and "Location" neglects the context of "Area" thereby diminishing the matching accuracy.

In this work, we propose a neural network approach based on pre-trained models to capture attribute matching information for deep entity matching (EM). To establish the correspondence between entity attributes, following the hierarchical structure Token→Attribute→Entity, we compare individual tokens within entities and across entities to obtain token similarity information. By subsequently aggregating the similarity information among tokens, we uncover complex attribute relationships in heterogeneous data. Ultimately, entity similarity is derived by evaluating the similarity of attribute values. To address matching challenges in heterogeneous data, we first learn contextual representations of tokens for a given pair of entities. Subsequently, within each entity, we leverage self-attention mechanisms to ascertain token dependencies, thus determining the significance of tokens within entities. This is followed by cross-entity token alignment using interaction attention mechanisms, yielding token similarity between entities. The aggregated token similarity is then weighted to derive attribute similarity. Concurrently, candidate entity pairs are serialized into sentence inputs for BERT model, generating sentence-level embeddings to mitigate the impact of data heterogeneity. Subsequently, within a Linear layer, heightened emphasis is placed on the matching degree of similar attributes, harnessing more attribute information while disregarding the influence of dissimilar attributes. This comprehensive approach culminates in the determination of entity matching outcomes.
Our main contributions can be summarized as follows:
\begin{itemize}
    \item [1)]
We employ BERT for contextual embeddings, enabling richer semantic and contextual information to be learned from a reduced dataset and producing more expressive token embeddings.
    \item [2)]
Building upon the transformation of entity matching into sequence pair classification, we introduce attribute similarity. This inclusion grants heightened focus to similar attributes, effectively harnessing entity attribute information.
    \item [3)]
We crawled data about Songhua River Basin and Liao River Basin from Wikipedia and Baidu Baike, resulting in a dataset encompassing 4039 reservoirs and 6576 river data entries. We constructed a water resources dataset and validated the model's effectiveness and robustness on this dataset.
\end{itemize}
\section{Related Work}
Entity Matching (EM), also known as entity resolution or record linkage, has been a subject of research in data integration literature for several decades\cite{mudgal2018deep}. To mitigate the high complexity of directly matching every pair of data, EM is typically divided into two steps: blocking and matching.

In recent years, matching techniques have garnered increased research attention. \cite{li2020deep}Ditto leverages pre-trained models such as BERT to transform entity matching into a binary classification problem of sequence pairs. It accomplishes this by inserting data attributes and values into special COL and VAL markers, concatenating them into sequence pairs, and then inputting them into the pre-trained model. This enables the model to classify sequence pairs and thus perform entity matching tasks.

DER-SSM\cite{sun2022towards} introduces and implements soft pattern matching, flexibly associating the relationships between attributes by considering inter-word correlations. It aggregates word information during entity matching to express relationships between attributes, greatly enhancing the effectiveness of entity matching for complex and corrupted data.

JointMatcher\cite{ye2022jointmatcher} employs relevance-aware encoders and numeral-aware encoders to enhance the model's focus on similar segments and numeral segments within sequences, thereby improving the accuracy of entity matching. HHG\cite{fu2021hierarchical} pioneers the use of graph neural networks to establish a hierarchical structure among words, attributes, and entities. By learning entity embeddings from top to bottom and capturing more contextual information, it enhances the derivation of entity embeddings.

Ditto utilizes the BERT model to transform entity matching into a binary classification problem of sequence pairs, better exploiting the contextual information of tokens. DER-SSM considers soft patterns to establish correspondences between attributes, mitigating the impact of heterogeneous data. Meanwhile, JointMatcher prioritizes the matching degree of similar segments between entities during the matching process. We have comprehensively considered these methods and, based on the foundation of using the BERT model to convert entity matching into binary classification of sequence pairs, we focus on the matching degree of similar attributes to address the issues of inadequate utilization of semantic information and matching of heterogeneous data.

\section{Preliminaries}
This section provides a formal definition of Entity Matching (EM) and subsequently outlines an LM-based approach to solving EM.
\subsection{Entity Matching}
\label{sec:title}
Entity Matching, also known as entity resolution, refers to the process of identifying pairs of records from structured datasets or text corpora that correspond to the same real-world entity. Let D be a collection of records from one or multiple sources, such as rows of relational tables, XML documents, or text paragraphs. Entity Matching typically involves two steps: blocking and matching . In this paper, we focus on the matching step of entity matching. Formally, we define the entity matching problem as follows:

Input: A set M of pairs of records to be matched.For each pair$(e_1,e_2)\in M_{}$, $e=\{(attr_i,\nu al_i)\}_{1\leq i\leq k}$each entity is represented in the form of K key-value pairs, where Key and Value are respectively the attribute name and attribute value of the entity.

Output: A set of pairs of records$M^{*}$, where each entity in each pair $(e_1,e_2)$points to the same entity, indicating the same real-world entity.

In this definition, our input is sufficiently general to apply to both structured and textual data. Attribute names and attribute values can take any form, including even indices or other identifiers, even if their true semantics are not available, such as "attr1" and "attr2".

\subsection{Methodology Framework}
\label{sec:title}
the schema of an entity represents an abstracted representation of the basic information about that entity. Schema matching is often a necessary prerequisite in the context of Entity Matching (EM), as there might be differences among attributes of different entities. Traditional EM methods typically establish one-to-one mapping relationships between attributes from different entities. However, in reality, the associations between two entity attributes can be intricate, and simple one-to-one mappings may fail to capture these complex relationships. To address EM more effectively, it becomes crucial to consider the intricate associations between attributes during the entity matching process, thereby enhancing the performance of entity matching.

For entities comprising distinct attributes, an entity itself can be seen as an instance of a schema. Given two entities,$e_1=\{<a_1^s,\nu_1^s>...<a_m^s,\nu_m^s>\}$ and $e_{2}=\{<a_{1}^{t},\nu_{1}^{t}>...<a_{n}^{t},\nu_{n}^{t}>\}$ , $\{a_1^s,a_2^s,...,a_m^s\}$ and $\{a_1^t,a_2^t,...,a_n^t\}$ respectively denote the distinct schemas of the two entities, and each schema is composed of Tokens representing attribute values of the entities.

\begin{figure} [ht]
   \begin{center}
   \begin{tabular}{c} 
   \includegraphics[height=5cm]{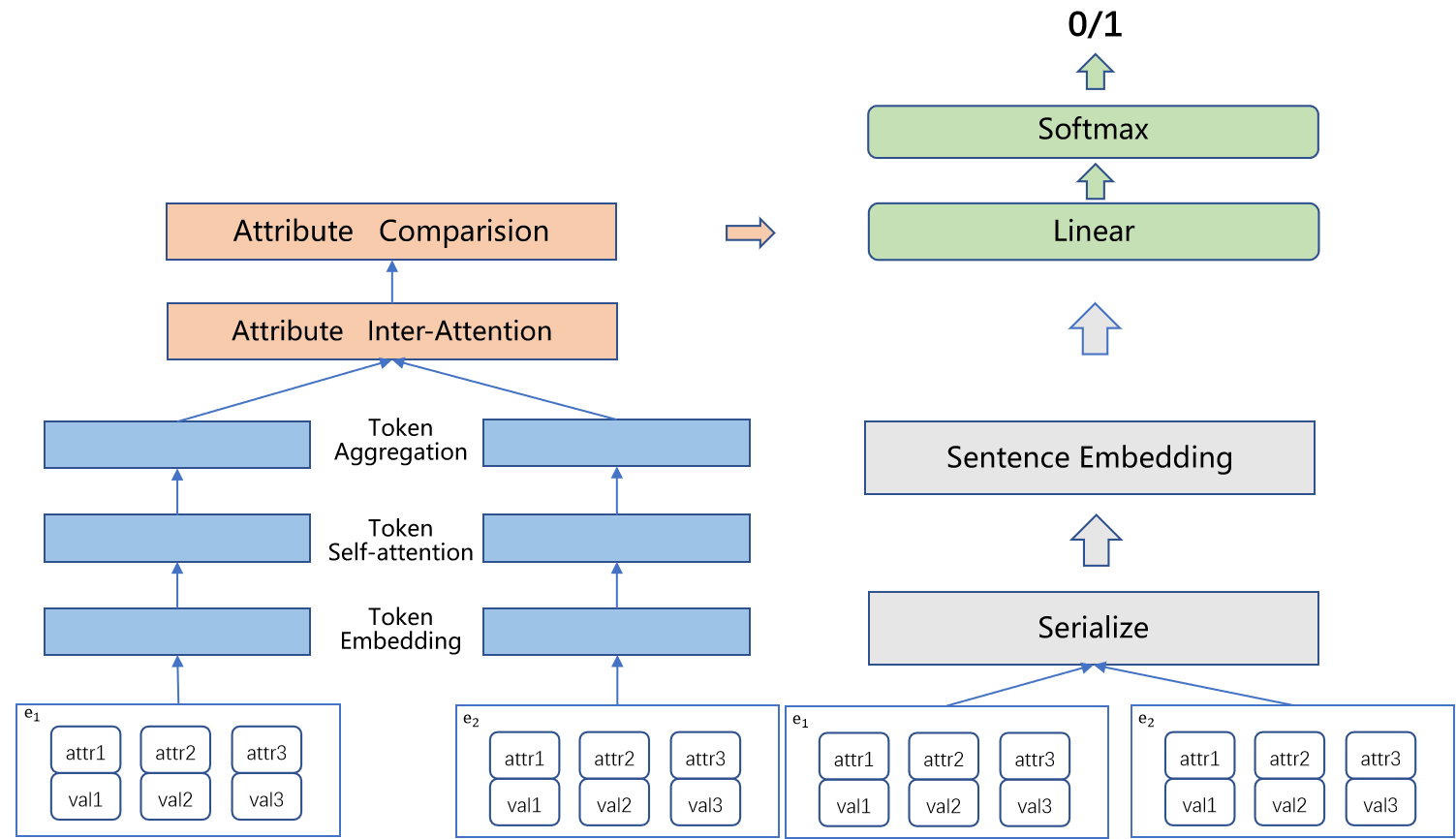}
   \end{tabular}
   \end{center}
   \caption[example] 
   { \label{fig:model} 
Model Architecture, On the left is how to capture complex relationships between attributes, and on the right is how to incorporate these complex relationships into the matching process.}
   \end{figure}

To achieve this goal, we construct a neural network based on BERT for entity matching. As illustrated in Fig.\ref{fig:model}, the left part involves a neural network that captures the complex associations between entity attributes. On the right side of the matching process, the final matching results are derived by considering the association matrix, which focuses on the degree of matching between different entity attributes. The matching process of the network mainly comprises the following steps:
(1) Token Embedding: Converting Tokens within attribute values into vectors using BERT, while capturing contextual relationships between each Token.
(2) Token Self-attention: Obtaining attention scores between Tokens of the same entity through self-attention mechanism.
(3) Token Aggregation: Aggregating attention scores between Tokens to obtain similarity information.
(4) Attribute Inter-Attention: Determining attention scores between Tokens of different entities through interactive attention.
(5) Attribute Comparison: Aggregating similarity scores between Tokens within and between entities to create a similarity matrix for attributes.
(6) Serialize: Serializing entities in the form of <Key, Value>.
(7) Sentence Embedding: Converting the serialized result into sentence vectors.
(8) Linear: Focusing on the matching degree of similar attributes within sentence vectors.
(9) Softmax: Normalizing the output of the Linear layer, resulting in a match (0/1) output.

\section{EM-CAR MODEL}
In this section, we will mainly introduce the specific implementation of each step of the model.

\subsection{Token Embedding}
Each token of both attribute values and attribute names needs to be embedded into a low-dimensional vector for subsequent calculations. Since attribute values are composed of sequences of different tokens, in the embedding process, the same token may have different vector representations in different contexts. Therefore, contextual semantic information should be integrated into the vectors. BERT is a pre-trained deep bidirectional Transformer model that effectively captures the semantic information of each token in its context through unsupervised learning on large-scale corpora. This implies that the token vectors generated by BERT can better represent the meaning of each token. Therefore, we choose to use the pre-trained BERT model for token embedding.
\textbf{\begin{equation}
\label{eq:fov}
H_i^s=BERT(A_i^sV_i^s)\
\end{equation}}
In the process of obtaining the attribute similarity matrix, the vector representation of entities is obtained by concatenating different tokens.$H^S=[H_1^s,H_2^s,...,H_m^s]$, $H^T=[H_1^s,H_2^s,...,H_n^t]$.

\subsection{Token Self-attention}

During the matching process, it is necessary to compare the similarity of each attribute, so we need to determine the significance of attributes within an entity. This is achieved through Token Self-attention, which computes the weights of tokens to aggregate the importance of attributes. Its role is to establish relationships between each token and other tokens within the sequence, capturing significant interdependencies to derive the importance of attributes. Through Self-attention, each token can be weighted and combined based on the importance of other tokens in the sequence, thereby better reflecting contextual information and semantic dependencies. Such an attention mechanism enables the model to dynamically focus on important tokens while disregarding less significant ones. Consequently, token-level self-attention is employed to weight the tokens within an entity. For attributes, their self-attention scores are computed using trainable matrices, as shown in the equation.
\textbf{\begin{equation}
\label{eq:fov}
\alpha_i^s=softmax((H_i^s)^TW_sH_i^s)\
\end{equation}}

\subsection{Token Aggregation}
To better harness token information for token-level comparisons, we employ Token Aggregation to merge the representations obtained after the Self-Attention operation, creating a comprehensive representation of the entire entity. This aggregation fuses all token information from the sequence into a single vector, enhancing the overall representation of the entity's information. This process involves an attention matrix, but we require a token weight vector for token aggregation. Therefore, we utilize a transformation function $m2\nu()$ to convert it into a token weight vector $W_{i}$ . $m2\nu()$  By summing the aggregation of each row of $\alpha_{i}^{s}$, a vector is derived, and subsequently, each element of the vector is normalized by dividing it by the maximum element.
\textbf{\begin{equation}
\label{eq:fov}
\alpha_i^{\prime s}=m2\nu(\alpha_i^s)\
\end{equation}}

\subsection{Attribute Inter-Attention}
The aforementioned operations yield attribute relationships within individual entities. However, our objective is to capture the complex relationships between attributes in heterogeneous data. Therefore, it is necessary to perform matching across different entities. Through Attribute Inter-Attention in entity matching tasks, interactive attention calculation is applied to different entity attributes. This process yields correspondences between attributes of different entities, enabling the learning of correlations among various attributes. As a result, a better grasp of the associations between entities is achieved. This attention mechanism aids in focusing on attributes relevant to matching while disregarding irrelevant ones. Each entity is considered as a sequence concatenated by tokens, inter-entity interaction attention is leveraged to obtain interaction representations between ${ H _ i ^ s}$ and ${ H _ t ^ j}$. Here, $W_{i\to T}$denotes the inter-entity interaction attention.
\textbf{\begin{equation}
\label{eq:fov}
{\beta^{i\to T}=softmax((\dot H_i^s)^TW_{i\to T}H^T)}
\end{equation}}
\textbf{\begin{equation}
\label{eq:fov}
\dot{H}_i^s=\beta^{i\to T}H^T\
\end{equation}}

\subsection{Attribute Comparision}

After obtaining the cross-entity correspondences, we need to use these correspondences to calculate the similarity between entity attributes, thus deriving the relationships between attributes. Attribute Comparison involves comparing different attributes of distinct entities during the matching process, computing their similarity or dissimilarity, and thereby gauging the level of association between different attributes. This attribute comparison mechanism aids the model in capturing essential features among attributes in entity matching tasks, further assisting the model in entity-level matching or classification. We apply element-wise absolute difference and Hadamard product to ${ H _ i ^ s}$ and$\dot{H}_i^s$ , and incorporate the intermediate representation into a highway network. Subsequently, the token-level similarity $\overset{s}{\operatorname*{C}}$ from $e_1$ to $e_2$ is the output of the highway network.
\textbf{\begin{equation}
\label{eq:fov}
C=HighwayNet([H^s-H^s],[H^s\otimes H^s])\
\end{equation}}
Lastly, the aggregation of token similarities obtained through the interaction attention mechanism of self-attention yields the similarity between entity attributes.
\textbf{\begin{equation}
\label{eq:fov}
\delta_i^s=\sum_{x\in[1,|H_i^5|]}C_i^s(x)\alpha_i^{\prime s}(x)\overset{s}{\operatorname*{C}}\
\end{equation}}

\textbf{\begin{equation}
\label{eq:fov}
R_{ij}^{S\to T}=\sum_{x\in[1,|H_{i}^{s}|]}C_{i}^{s}(x)\alpha_{i}^{\prime{s}}(x)\overset{s}{\operatorname*{C}}\
\end{equation}}

\subsection{Serialize and Sentence-Embedding}

We employ the methods from Ditto \cite{li2020deep} to serialize the data and generate sentence embeddings.
For each entity pair, we serialize it as follows:
$serialize(e)=[COL]attr_1[VAL]val_1[COL]attr_2[VAL]val_2...[COL]attr_k[VAL]val_k$ ,
Where [COL] and [VAL] are special tokens used to indicate the start of attribute names and values, respectively. For example, the first entry in the second table is serialized as:
For each candidate entity pair,
$serialize(e_1,e_2)=[CLS]serialize(e_1)[SEP]serialize(e_2)[CLS]$,
where [SEP] is a special token that separates the two sequences, and [CLS] is a special token required by BERT to encode the sequence pair into a 768-dimensional vector.

\subsection{Linear and Softmax}
In entity matching tasks, a linear layer is employed to perform a linear transformation on the vectors that have undergone feature extraction and encoding. The formula representing the linear layer for input feature vector X is as follows:
\begin{equation}
\label{eq:fov}
L(X,W,b)=X\bullet W+b\
\end{equation}
Here, W represents the weight matrix to be learned, and b is the bias vector. In this context, the vector X is obtained through previous serialization and sentence embedding, resulting in the embedded vector E for entity pairs. The matrix W corresponds to the obtained entity attribute similarity matrix

Further applying a softmax function yields the output vector. This vector represents a probability distribution, where each element signifies the probability for the respective category. In this context, the output of 0 signifies a non-match, while 1 signifies a match.
\section{Experiment}
In this section, we utilized a dataset to evaluate our EM-CAR model.
\subsection{Experiment Dataset}
When evaluating our approach, we utilized various types of datasets, including:
Two isomorphic public datasets\cite{konda2016magellan} (simple 1:1 attribute associations).
Two heterogeneous public datasets (complex associations of 1:m and m:n).
A hydraulic heterogeneous dataset (complex associations of 1:m and m:n).
The information for the public datasets is presented in the following table.
\begin{table}[ht]
\caption{The "Size" column indicates the size of the "Size" table, "\#POS." represents the number of positive matches, and "\#ATTR." represents the attribute number. The attribute association "m:n" between two patterns is entirely different from the attribute numbering "c-d". The "m:n" attribute association signifies the presence of at least one complex 1:m or m:n attribute association between two patterns. The attribute numbering "c-d" only denotes that the first pattern has "c" attributes, while the second pattern has "d" attributes.} 
\label{tab:fonts}
\begin{center}       
\begin{tabular}{|l|l|l|l|l|l|} 
\hline
\rule[-1ex]{0pt}{3.5ex}  Type & Dataset & Domain & Size & \#POS. & \#ATTR.  \\
\hline
\rule[-1ex]{0pt}{3.5ex}  Same pattern & iTunes-Amazon & Music & 539 & 132 & 8-8   \\
\hline
\rule[-1ex]{0pt}{3.5ex}   Same pattern & DBLP-Scholar & Citation & 28707 & 5347 & 4-4   \\
\hline
\rule[-1ex]{0pt}{3.5ex}  Different pattern & UIS1-UIS2 & Person & 12853 & 2736	& 4-4   \\
\hline
\rule[-1ex]{0pt}{3.5ex}  Different pattern & Walmart-Amazon	& Electronics & 10242 & 962 & 4-4   \\
\hline
\end{tabular}
\end{center}
\end{table} 
Homogeneous Data: The patterns of homogeneous data involve simple associations (1:1). iTunes-Amazon (iA) and DBLP-Schoral1 (DS1) correspond respectively to iTunes-Amazon1 and DBLP-choral1\cite{mudgal2018deep}.

Heterogeneous Data: The complex data SM involves complex associations (1:m or m:n). We implemented a variant of the Synthetic Data Generator UIS to generate the UIS1-UIS2 (UU) dataset\cite{thirumuruganathan2018reuse}. The initial five attributes are name, address, city, state, and zip code. Address and city are combined into a new attribute in UIS1 records, while city and state are integrated into a new attribute in UIS2 records. Therefore, the attribute numbering for UU is 4–4. Walmart-Amazon1 (WA1) is a variant of Walmart-Amazon (5-5)\cite{jurek2017novel} . Brand and model are merged into a new attribute in Walmart records, while category and model are integrated into a new attribute in Amazon records. Hence, the attribute numbering for WA1 is 4–4.

Next, we introduce the composition of the hydraulic dataset, which we crawled separately from Wikipedia and Baidu Baike. We collected data about the Songhua River Basin and the Liao River Basin, including 4039 reservoirs and 6576 river records. Reservoirs include attributes such as reservoir name, location, area, region, normal water level, watershed area, normal storage level, etc. Rivers include attributes such as river name, region, river grade, river length, basin area, etc. As shown in Figure 1, there exist complex correspondences among these attributes. During data processing, we labeled data belonging to the same entity as matching and labeled data pointing to different entities as non-matching. Since this data was crawled from web pages based on names, most of the non-matching entities are entities with the same name, and the proportion of same-name entities in the data is not high. Therefore, there are not enough negative examples in the data. We created some negative examples by replacing attribute values with synonyms. Finally, our dataset has a size of 21230, including 5000 positive instances, with 2000 positive instances for reservoirs and 3000 for rivers.

\subsection{Implementation and Setup}
We implemented our model using PyTorch and the Transformers library. In all experiments, we used the base uncased variants of each model. We further applied mixed-precision training (fp16) optimization to accelerate both training and inference speed. For all experiments, we fixed the maximum sequence length to 256, set the learning rate to 3e-5, and employed a linear decay learning rate schedule. The training process runs for a fixed number of epochs (10, 15, or 40 depending on the dataset size) and returns the checkpoint with the highest F1 score on the validation set.

Comparison Methods. We compare EM-CAR with state-of-the-art EM solutions such as Ditto, the attribute correspondence-aware method DEM-SSR, and the classical method DeepMatcher.
Here's a summary of these methods. We report the average F1 score over 6 repeated runs in all settings.

DeepMatcher: DeepMatcher is a state-of-the-art classical method , customizes an RNN architecture to aggregate attribute values and then compare/align the aggregated representations of attributes. 

Ditto: Ditto is a state-of-the-art matching solution that employs all three optimizations, Domain Knowledge (DK), TF-IDF summarization (SU), and Data Augmentation (DA).

DER-SSM: In comparison to Ditto, DER-SSM defines and implements soft pattern matching, obtaining context relations between tokens through BiGRU. It considers soft pattern matching by aggregating token similarity during entity matching based on the context relationships between tokens.

\subsection{Experiment Result}

The F1 score is used to measure the precision of entity matching (EM) and is the harmonic mean of precision (P) and recall (R). Precision (P) represents the score of correct matching predictions, while recall (R) represents the score of true matches predicted as matches.

Typically, in EM, there are two phases: blocking and matching\cite{li2020deep} . Our focus is on the matching phase in entity matching (EM), assuming that blocking has already been performed. We follow the same blocking setup\cite{li2020deep} , where blocking is applied to generate a candidate set for the dataset. All pairs in the candidate set are labeled. The dataset is then divided into a 3:1:1 ratio for training, validation, and testing.
\begin{table}[ht]
\caption{Average F1 Scores of Different Methods.} 
\label{tab:fonts}
\begin{center}       
\begin{tabular}{|l|l|l|l|l|l|} 
\hline
\rule[-1ex]{0pt}{3.5ex}  Type & Dateset & DeepMatcher & DER-SSM & Ditto & EM-CAR \\
\hline
\rule[-1ex]{0pt}{3.5ex}   Same pattern & iTunes-Amazon & 82.3 & 85.7 & \textcolor{blue}{89.6} & 89.5 \\ 
\hline
\rule[-1ex]{0pt}{3.5ex}   Same pattern & DBLP-Scholar & 85.4 & 89.2 & 90.1 & \textcolor{blue}{90.9} \\
\hline
\rule[-1ex]{0pt}{3.5ex}  Different pattern & UIS1-UIS2 & 76.2 & 80.4 & 85.2 & \textcolor{blue}{86.3} \\
\hline
\rule[-1ex]{0pt}{3.5ex}  Different pattern & Walmart-Amazon & 77.1 & 81.2 & 84.4 & \textcolor{blue}{85.3} \\
\hline
\rule[-1ex]{0pt}{3.5ex}  Water data & SongLiao & 74.3 & 78.2 & 80.1 & \textcolor{blue}{81.2} \\
\hline
\end{tabular}
\end{center}
\end{table} 

To further demonstrate the performance of EM-CAR, we conducted a case study comparing it with Ditto. First, it should be noted that Ditto directly utilizes context-based embeddings obtained from pre-trained language models (LM) for classification, making it not entirely suited for entity matching (EM) tasks. Specifically,Ditto's embeddings might not be fully optimized for the specific task of entity matching, as they are derived from a broader range of language modeling objectives. This could potentially limit Ditto's ability to capture the nuanced and complex relationships between attributes required for accurate entity matching.

In contrast, our model aims to address this issue by placing greater emphasis on attributes with higher similarity. As depicted in the figure, our model takes into account the similarity of attributes, particularly those that are more closely related, This approach allows our model to better capture the nuanced relationships between attributes and improve the overall matching accuracy.

   \begin{figure} [ht]
   \begin{center}
   \begin{tabular}{c} 
   \includegraphics[width=\textwidth,height=4cm]{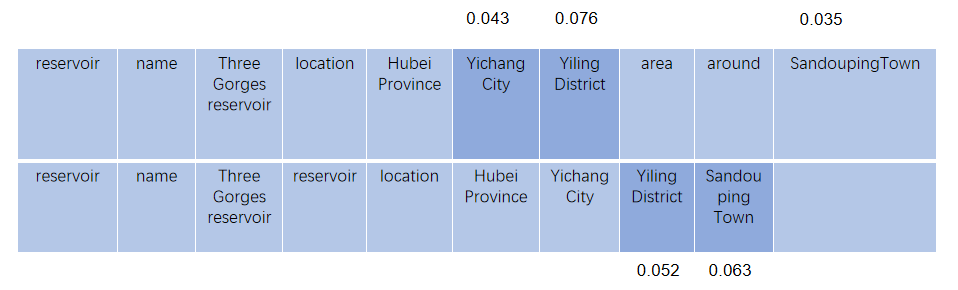}
   \end{tabular}
   \end{center}
   \caption[error] 
   { \label{fig:error} 
Attention scores of Ditto.}
   \end{figure} 
    \begin{figure} [ht]
   \begin{center}
   \begin{tabular}{c} 
   \includegraphics[width=\textwidth,height=4cm]{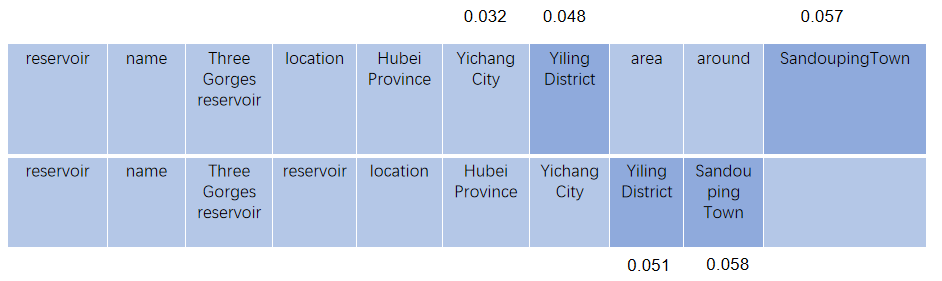}
   \end{tabular}
   \end{center}
   \caption[true] 
   { \label{fig:true} 
Attention scores of EM-CAR.}
   \end{figure} 
As shown in the Fig.\ref{fig:error}, when performing matching, Ditto's utilization of pre-trained models can lead to erroneous judgments. This is attributed to the fact that, while determining whether these two entities match, the top two attention scores are placed on the tokens "YichangCity" and "YilingDistrict" while the score for "SandoupingTown" is not as high. Consequently, more attention is directed towards the correspondence between "YichangCity" and "YilingDistrict" as well as "YilingDistrict" and "SandoupingTown" .

In contrast, we aim to prioritize the matching probability between "YilingDistrict" and "SandoupingTown" As illustrated in Fig.\ref{fig:true}, our model focuses more on the matching degree of similar attributes, denoted by $(e_1, e_2)$. This enables our model to appropriately emphasize the similarity between attributes and achieve accurate results.

%

\section{Conclusion}
In this paper, We propose EM-CAR, a method that leverages attribute similarity within the context of pre-trained models to address complex entity correspondence. In our approach, we compare the classical DeepMatcher, DER-SSM (which considers soft patterns, i.e., complex attribute correspondences), and Ditto, which employs pre-trained models. We evaluate these methods, including our own, on three types of datasets: homogeneous, heterogeneous, and hydraulic data (heterogeneous). Across all datasets, DeepMatcher achieves the lowest F1 score due to its reliance on a simple CNN network, which struggles to capture semantic information effectively.

For the two homogeneous public datasets, DER-SSM and Ditto exhibit comparable accuracy, as shown in the figure. Homogeneous datasets feature straightforward 1:1 relationships, thus methodological differences have less pronounced impacts. The primary distinction lies in whether a pre-trained model is utilized.

Concerning the two heterogeneous public datasets, DER-SSM initially shows promise, but its use of BIGRU limits semantic context to a local n-character window, resulting in slightly lower accuracy compared to Ditto. In contrast, our model takes into account complex attribute correspondences, placing greater emphasis on matching similar attributes, thereby enhancing accuracy to a certain extent.

On the hydraulic dataset, limited training data affects all three models' performance, resulting in reduced accuracy. However, our model still achieves the highest F1 score. This suggests that prioritizing the matching of similar attributes during the matching process has a positive impact on improving matching accuracy.

In summary, our EM-CAR approach effectively enhances entity matching accuracy by focusing on the similarity between attributes, especially for complex correspondences, as demonstrated across various datasets in comparison to other methods such as DeepMatcher, DER-SSM, and Ditto.

\vspace{2cm}

\section*{Authors}
\noindent {\bf Jiamin. Lu }  Assistant Professor at Information Department in Hohai University, China. He received his Ph.D degree in Information Science from FernUniversität in Hagen, Germany, 2014. His research interests include parallel processing on MOD (Moving Object Database), data management in Knowledge Graph construction.  At present, he is mainly working on the conjunction of big data technologies and smart water applications.\\

\noindent {\bf Shitao. Wang} He is pursuing a master's degree in Computer Science at Hohai University. His research interests include knowledge graph, entity matching, and natural language processing..\\

\end{document}